\documentclass{article} 
\usepackage{nips13submit_e,times}
\usepackage{hyperref}
\usepackage{url}
\usepackage{bm}
\usepackage{latexsym}
\usepackage{amsmath}
\usepackage{amssymb}
\usepackage{booktabs}
\usepackage{epsfig}
\usepackage{graphicx}
\usepackage{subfigure}
\usepackage{xspace}

\title{One-Shot Adaptation of Supervised Deep Convolutional Models}

\author{
Judy Hoffman, Eric Tzeng, Jeff Donahue \\
UC Berkeley, EECS \& ICSI\\
\footnotesize{\texttt{\{jhoffman,etzeng,jdonahue\}@eecs.berkeley.edu} }\\
\And
Yangqing Jia\thanks{This work was completed while Yangqing Jia was a graduate student at UC Berkeley} \\
Google Research \\
\texttt{jiayq@google.com} \\
\AND
Kate Saenko \\
UMass Lowell, CS \& ICSI \\
\footnotesize{\texttt{saenko@cs.uml.edu}} \\
\And
Trevor Darrell \\
UC Berkeley, EECS \& ICSI\\
\footnotesize{\texttt{trevor@eecs.berkeley.edu}} \\
}

\newcommand{\daume}{Daum\'e~III\xspace}

\nipsfinalcopy 

\begin{document}

\maketitle

\begin{abstract}
Dataset bias remains a significant barrier towards solving real world computer vision tasks.
Though deep convolutional networks have proven to be a competitive approach for image classification, a question remains: have these models have solved the dataset bias problem?
In general, training or fine-tuning a state-of-the-art deep model on a new domain requires a significant amount of data, which for many applications is simply not available.
Transfer of models directly to new domains without adaptation has historically led to poor recognition performance.
In this paper, we pose the following question: is a single image dataset, much larger than previously explored for adaptation, comprehensive enough to learn general deep models that may be effectively applied to new image domains? In other words, are deep CNNs trained on large amounts of labeled data as susceptible to dataset bias as previous methods have been shown to be?
We show that a generic supervised deep CNN model trained on a large dataset reduces, but does not remove, dataset bias.
Furthermore, we propose several methods for adaptation with deep models that are able to operate with little (one example per category) or no labeled domain specific data.
Our experiments show that adaptation of deep models on benchmark visual domain adaptation datasets can provide a significant performance boost.

\end{abstract}

\section{Introduction}

Supervised deep convolutional neural networks (CNNs) trained on large-scale
classification tasks have been shown to learn impressive mid-level structures
and obtain high levels of performance on contemporary classification
challenges \cite{ilsvrc2012,zeiler-arxiv-2013}. These models generally assume
extensive training using labeled data, and testing is limited to data from the
same domain. In practice, however, the images we would like to classify are
often produced under different imaging conditions or drawn from a different
distribution, leading to a domain shift. Scaling such models to new domains
remains an open challenge.

Deep CNNs require large amounts of training data to learn
good mid-level convolutional models and final fully-connected classifier
stages. While the continuing expansion of web-based datasets like
ImageNet \cite{ilsvrc2012} promises to produce labeled data for almost any desired
category, such large-scale supervised datasets
may not include images of the category across all
domains of practical interest. Earlier deep learning efforts addressed this
challenge by learning layers in an unsupervised fashion using unlabeled data to
discover salient mid-level structures \cite{coates-nips12, dean-nips12}. While such approaches are appealing, they
have heretofore been unable to match the level of performance of supervised
models, and unsupervised training of networks with the same level of depth
as \cite{supervision} remains a challenge.

Unfortunately, image datasets are inherently biased \cite{efros-cvpr11}. 
Theoretical \cite{ben2007analysis, blitzer2007learning} and practical results from \cite{saenko-eccv10,efros-cvpr11} have shown that supervised methods' test error increases in proportion to the difference between the test and training input distribution. 
Many visual domain adaptation methods have been put forth to compensate for dataset bias \cite{daume,yang-icdm07,aytar-iccv11,saenko-eccv10,kulis-cvpr11,Khosla-eccv12,gopalan-iccv11,gong-cvpr12,hoffman-eccv12,hoffman-iclr13}, but are limited to shallow models. 
Evaluation for image category classification across visually distinct domains has focused on the Office dataset, which contains 31 image categories and 3 domains \cite{saenko-eccv10}. 
Recently, \cite{deeplearning-arxiv-2013} showed that using the deep mid-level features learned on ImageNet, instead of the more conventional bag-of-words features, effectively removed the bias in some of the domain adaptation settings in the Office dataset \cite{saenko-eccv10}.
However, \cite{deeplearning-arxiv-2013} limited their experiments to
small-scale source domains found only in Office, and evaluated on only a subset
of relevant layers.

Yet until now, almost none of the previous domain adaptation studies used ImageNet as the \textit{source} domain, nor utilized the full set of parameters of a deep CNN trained on source data. Recent work by Rodner et al.~\cite{rodner-arxiv13} attempted to adapt from ImageNet to the SUN dataset, but did not take advantage of deep convolutional features. 

In this paper, we ask the question: will deep models still suffer from dataset
bias when trained with all layers of the CNN and a truly large scale source
dataset?
Here, we provide the first evaluation of domain adaptation
with deep learned representations in its most natural setting, in which all of
ImageNet is used as source data for a target category.  We use the 1.2 million
labeled images available in the 2012 ImageNet 1000-way classification
dataset~\cite{ilsvrc2012} to train the model in \cite{supervision} and evaluate
its generalization to the Office dataset. This constitutes a three
orders of magnitude increase in source data compared to the several thousand
images available for the largest domain in Office.

We find that it is easier to adapt from ImageNet than from previous smaller source domains, but  that dataset bias remains a major issue. Fine-tuning the parameters on the small amount of labeled target data (we consider one-shot adaptation) turns out to be unsurprisingly problematic. Instead, we propose a simple yet intuitive adaptation method: train a final domain-adapted classification ``layer'' using various layers of the pre-trained network as features, without any fine-tuning its parameters.
We provide a comprehensive evaluation of existing methods for classifier adaptation as applied to each of the fully connected layers of the network, including the last, task-specific classification layer.
When adapting from ImageNet to Office, it turns out to be possible to achieve target domain performance on par with source domain performance using only a single labeled example per target category. 

We examine both the setting where there are a few labeled examples from the target domain (\emph{supervised adaptation}) and the setting where there are no labeled target examples (\emph{unsupervised adaptation}). We also describe practical solutions for choosing between the various adaptation methods based on experimental constraints such as limited computation time. 

\vspace{-.3cm}
\section{Background: Deep Domain Adaptation Approaches}
\vspace{-.2cm}
\label{sec:decaf}
For our task we consider adapting between a large source domain and a target domain with few or or no labeled examples. 
A typical approach to domain adaptation or transfer learning with deep architectures is to take the representation learned via back-propagation on a large dataset, and then transfer the representation to a smaller dataset by fine-tuning, i.e. backpropagation at a lower learning rate \cite{rcnn,zeiler-arxiv-2013}.
However, fine-tuning requires an ample amount of labeled target data and so should not be expected to work well when we consider the very sparse label condition, such as the \textit{one-shot learning} scenario we evaluate below, where we have just one labeled example per category in the target domain.

In fact, in our experiments under this setting, fine-tuning actually reduces performance.
Specifically, on the ImageNet$\rightarrow$Webcam task reported in Section~\ref{sec:eval}, using the final output layer as a predictor in the target domain received 66\% accuracy, while using the final output layer after fine tuning produced a degraded accuracy of 61\%.

A separate method that was recently proposed for deep adaptation is called Deep Learning for domain adaptation by Interpolating between Domains (DLID)~\cite{ref:dlid}. This method learns multiple unsupervised deep models directly on the source, target, and combined datasets and uses a representation which is the concatenation of the outputs of each model as its adaptation approach. 
While this was shown to be an interesting approach, it is limited by its use of unsupervised deep structures. 

In general, unsupervised deep convolutional models have been unable to achieve the performance of supervised deep CNNs. However, training a supervised deep model requires sufficient labeled data. Our insight is that the extensive labeled data available in the source domain can be exploited using a supervised model without requiring a significant amount of labeled target data.

Therefore, we propose using a supervised deep source model with supervised or unsupervised adaptation algorithms that are applied to models learned on the target data directly. 
This hybrid approach will utilize the strong representation available from the supervised deep model trained on a large source dataset while requiring only enough target labeled data to train a shallow model with far fewer parameters. 
Specifically, we consider training a convolutional neural network (CNN) on the source domain and using that network to extract features on the target data that can then be used to train an auxiliary shallow learner. 
For extracting features from the deep source model, we follow the setup of Donahue et al.~\cite{deeplearning-arxiv-2013}, which extracts a visual feature \textit{DeCAF} from the ImageNet-trained architecture of~\cite{supervision}.

\vspace{-.3cm}
\section{Adapting Deep CNNs with Few Labeled Target Examples}
\label{sec:adapt-algs}
\vspace{-.2cm}

\newcommand{\svmT}{\bm{\theta}}
\newcommand{\svmB}{b}
\newcommand{\svmAug}{\tilde{\svmT}}
\newcommand{\svmAugAll}{\bm{\Theta}}


We propose a general framework for selectively adapting the parameters of a convolutional neural network (CNN) whose representation and classifier weights are trained on a large-scale source domain, such as ImageNet.
Our framework adds a final domain-adaptive classification ``layer'' that takes the activations of one of the existing network's layers as input features. Note that the network cannot be effectively fine-tuned without access to more labeled target data. This adapted layer is a linear classifier that combines source and target training data using an adaptation method. To demonstrate the generality of our framework, we select a representative set of popular linear classifier adaptation approaches that we empirically evaluate in Section~\ref{sec:eval}. We separate our discussion into the set of supervised and unsupervised adaptation settings.

Below we denote the features extracted over the source domain as $\bm{X}$ and the features extracted over the target domain as $\tilde{\bm{X}}$. Similarly, we denote the source domain image classifier as $\bm{\theta}$ and the target domain image classifier as $\bm{\tilde{\theta}}$.

\subsection{Unsupervised Adaptation}
Many unsupervised adaptation techniques seek to minimize the distance between subspaces that represent the source and target domains. We denote these subspaces as $U$ and $\tilde{U}$, respectively.

{\bf GFK~\cite{gong-cvpr12}}
The Geodesic Flow Kernel (GFK) method~\cite{gong-cvpr12} is an unsupervised domain adaptation approach which seeks embeddings for the source and target points that minimize domain shift.
Inputs to the method are $U$ and $\tilde{U}$, lower-dimensional embeddings of the source and target domains (e.g. from principal component analysis).
The method constructs the geodesic flow $\phi(t)$ along the manifold of subspaces such that $U = \phi(0)$ and $\tilde{U} = \phi(1)$.
Finally, a transformation $G$ is constructed by computing
$
G =
\int_0^1 \phi(t) \phi(t)^{\intercal} dt
$
using a closed-form solution, and classification is performed by training an SVM on the source data $\bm{X}$ and transformed target data $G\tilde{\bm{X}}$.
\\\\
{\bf SA~\cite{sa}}
The Subspace Alignment (SA) method~\cite{sa} also begins with low-dimensional embeddings of the source and target domains $U$ and $\tilde{U}$, respectively.
It seeks to minimize in $M$, a transformation matrix, the objective
$ \| UM - \tilde{U} \|^2_F$.
The analytical solution to this objective is $M^* = U^{\intercal} \tilde{U}$.
Given $M^*$, an SVM is trained on source data $\bm{X}$ and transformed target data $U M^* \tilde{U}^{\intercal} \tilde{\bm{X}}$.

\subsection{Supervised Adaptation}

{\bf Late Fusion}
Perhaps the simplest supervised adaptation method is to independently train a source and target classifier and combine the scores of the two to create a final scoring function. 
We call this approach Late Fusion. It has been explored by many for a simple adaptation approach.
Let us denote the score from the source classifier as $v_s$ and the score from the target classifier as $v_t$.  For our experiments we explore two methods of combining these scores, which are described below:
\begin{itemize}
\item {\bf Max}: Produce the scores of both the source and target classifier and simply choose the max of the two as the final score for each example. Therefore, $v_{\text{adapt}} = \max(v_s, v_t)$.
\item {\bf Linear Interpolation}: Set the score for a particular example to equal the convex combination of the source and target classifier scores,  $v_{\text{adapt}} = (1-\alpha)v_{s} + \alpha v_{t}$. This method requires setting a hyperparameter, $\alpha$, which determines the weights of the source and target classifiers.
\end{itemize}

Late Fusion has two major advantages: it is easy to implement, and the source classifier it uses may be precomputed to make adaptation very fast. In the case of the linear interpolation combination rule, however, this method can potentially suffer from having a sensitive hyperparameter. We show a hyperparameter analysis in Section~\ref{sec:eval}.

{\bf \daume~\cite{daume}}
This simple feature replication method was proposed for domain adaptation by~\cite{daume}.
The method augments feature vectors with a source component, a target component, and a shared component.
Each source data point $\bm{x}$ is augmented to $\bm{x}' = (\bm{x}; \bm{x}; \bm{0})$,
and each target data point $\tilde{\bm{x}}$ is augmented to $\tilde{\bm{x}}' = (\tilde{\bm{x}}; \bm{0}; \tilde{\bm{x}})$.
Finally, an SVM is trained on the augmented source and target data---a relatively expensive procedure given the potentially large size of the source domain and the tripled augmented feature dimensionality.

{\bf PMT~\cite{aytar-iccv11}} 
This classifier adaptation method, Projective Model Transfer (PMT), proposed by~\cite{aytar-iccv11}, is a variant of adaptive SVM. It takes as input a classifier $\svmT$ pre-trained on the source domain.
PMT-SVM learns a target domain classifier $\svmAug$ by adding an extra term to the usual SVM objective which regularizes the angle
$
\alpha(\svmAug, \svmT) =
\cos^{-1}\left(
  \frac{
    \svmT^{\intercal} \svmAug
  }{
    \| \svmT \| \| \svmAug \|
  }
\right)
$
between the target and source hyperplanes.
This results in the following loss function:
\begin{align}
\mathcal{L}_{PMT}(\svmAug)
&=
\frac{1}{2} \| \svmAug \|^2_2
+
\frac{\Gamma}{2} \| \svmAug \|^2_2 \sin^2 \alpha(\svmAug, \svmT)
+
\ell_{hinge}(\tilde{\bm{X}}, \tilde{\bm{Y}}; \svmAug)
\enspace,
\end{align}
where
$
\ell_{hinge}(\bm{X}, \bm{Y}; \svmT)
$
denotes the SVM hinge loss of a data matrix
$\bm{X}$,
label vector
$\bm{Y}$,
and classifier hyperplane $\svmT$,
and $\Gamma$ is a hyperparameter which, as it increases, enforces more transfer from the source classifier.

{\bf MMDT~\cite{hoffman-iclr13}}
\newcommand{\mat}[1]{#1}
\newcommand\datransform{\mat{A}}
The Max-margin Domain Transforms (MMDT) method from~\cite{hoffman-iclr13} jointly optimizes an SVM-like objective over a feature transformation matrix $\datransform$ mapping target points to the source feature space and classifier parameters $\svmT$ in the source feature space.
In particular, MMDT minimizes the following loss function (assuming a binary classification task to simplify notation, and with
$\ell_{hinge}$
defined as in PMT):
\begin{align}
\mathcal{L}_{MMDT}(\svmT, \datransform)
&=
\frac{1}{2} \| \svmT \|^2_2
+
\frac{1}{2} \| \datransform - \mat{I} \|^2_F
+
C_s
\ell_{hinge}(\bm{X}, \bm{Y}; \svmT)
+
C_t
\ell_{hinge}(\datransform \tilde{\bm{X}}, \tilde{\bm{Y}}; \svmT)
\enspace,
\end{align}
where $C_s$ and $C_t$ are hyperparameters controlling the importance of correctly classifying the source and target points (respectively).

\section{Evaluation}
\label{sec:eval}



\subsection{Datasets}
The Office~\cite{saenko-eccv10} dataset is a collection of images from three
distinct domains: Amazon, DSLR, and Webcam. The 31 categories in the dataset
consist of objects commonly encountered in office settings, such as keyboards,
file cabinets, and laptops. Of these 31 categories, 16 overlap with the
categories present in the 1000-category ImageNet classification task\footnote{
The 16 overlapping categories are
\textit{backpack},
 \textit{bike helmet},
 \textit{bottle},
 \textit{desk lamp},
 \textit{desktop computer},
 \textit{file cabinet},
 \textit{keyboard},
 \textit{laptop computer},
 \textit{mobile phone},
 \textit{mouse},
 \textit{printer},
 \textit{projector},
 \textit{ring binder},
 \textit{ruler},
 \textit{speaker},
 and
 \textit{trash can}.
}.
Thus, for our experiments, we limit ourselves to these
16 classes.  In our experiments using Amazon as a source domain,
we follow the standard training protocol for this dataset of using 20 source
examples per category~\cite{saenko-eccv10,gong-cvpr12}, for a total of 320
images.

ImageNet~\cite{ilsvrc2012} is the largest available dataset of image category labels. We use 1000 categories' worth of data (1.2M images) to train the network, and use the 16 categories that overlap with Office (approximately 1200 examples per category or  $\approx$20K images total) as labeled source classifier data.

\subsection{Experimental Setup \& Baselines}

For our experiments, we use the fully trained deep CNN model described in Section~\ref{sec:decaf}, extracting feature
representations from three different layers of the CNN. We then train a source classifier using these features on one of two source domains, and adapt to the target domain.

The source domains we consider are either the Amazon domain, or the corresponding 16-category ImageNet subset where each category has many more examples.
We focus on the Webcam domain as our target (test) domain, as Amazon-to-Webcam  was shown to be the only challenging shift in \cite{deeplearning-arxiv-2013} (the DSLR domain is much more similar to Webcam and did not require adaptation when using deep mid-level features). This combination exemplifies the shift from online web images to real-world images taken in typical office/home environments.
Note that, regardless of the source domain chosen to learn the classifier, ImageNet data from all 1000 categories was used to train the network. 

In addition, for the supervised adaptation setting we assume access to only a single example per category from the target domain (Webcam). 

Each method is then evaluated across 20 random train/test splits, and we report averages and standard errors for each setting. 
For each random train/test split we choose one example for training and 10 other examples for testing (so there is a balanced test set across categories). Therefore, each test split has 160 examples. The unsupervised adaptation methods operate in a transductive setting, so the target subspaces are learned from the unlabeled test data.

\paragraph{Non-adaptive Baselines}
In addition to the adaptation methods outlined in Section~\ref{sec:adapt-algs},
we also evaluate using the following non-adaptive baselines.

\begin{itemize}
  \item{\textbf{SVM (source only)}: A support vector machine trained only on
    source data. 
}
  \item{\textbf{SVM (target only)}: A support vector machine trained only on
    target data.}
  \item{\textbf{SVM (source and target)}: A support vector machine trained on
  both source and target data. To account for the large discrepancy between the
  number of training data points in the source and target domains, we weighted
  the data points such that the constraints from the source and target domains
  effectively contribute equally to the optimization problem.  Specifically,
  each source data point receives a weight of $\frac{n_t}{n_s+n_t}$, and each target
  data point receives a weight of $\frac{n_s}{n_s+n_t}$, where $n_s,n_t$ denote the
  number of data points in the source and target, respectively.}
\end{itemize}

Many of the adaptation methods we evaluate have hyperparameters that must be
cross-validated for use in practice, so we set the parameters of the adaptation
techniques as follows.

First, the C value used for C-SVM in the classifier for all methods is set to $C=1$. Without any validation data we are not able to tune this parameter properly, so we choose to leave it as the default value. Since all methods we report require setting of this parameter, we feel that the relative comparisons between methods is sound even if the absolute numbers could be improved with a new setting for C.
  For \daume and MMDT, which look at the source and target data
  simultaneously, we use the same weighting scheme as we did for the source and
  target SVM.
  Late Fusion with the linear interpolation combination rule is reported across hyperparameter settings in Figure~\ref{fig:linint-eval} to help understand how performance varies as we trade off emphasis between the learned classifiers from the source and target domains. Again, we do not have the validation data to tune this parameter so we report in the tables the performance averaged across parameter settings. The plot vs $\alpha$ indicates that there is usually a best parameter setting that could be learned with more available data.
  For PMT, we choose $\Gamma=1000$, which corresponds to allowing a large amount of transfer from the source classifier to the target classifier. We do this because the source-only classifier is
  stronger than the target-only classifier (with ImageNet source). 
  For the unsupervised methods GFK and SA, again we evaluated a variety of
  subspace dimensionalities and Figure~\ref{fig:sagfk-eval} shows that the overall method performance does not vary significantly with the dimensionality choice.

\begin{table*}
\centering
\begin{tabular}{lccc}
\toprule
Adaptation Method & Training Data & DeCAF$_6$ & DeCAF$_7$ \\
\midrule
SVM (source only) & Amazon & $50.28 \pm 1.8$ & \textcolor{blue}{$\bm{54.08 \pm 1.7}$} \\
SVM (target only) & Webcam & $62.28 \pm 1.8$ & $64.97 \pm 1.8$ \\
\midrule
GFK \cite{gong-cvpr12} & Amazon & \textcolor{blue}{$\bm{53.13 \pm 1.1}$} & \textcolor{blue}{$\bm{53.39 \pm 1.1}$} \\
SA \cite{sa} & Amazon & $51.74 \pm 1.2$ & \textcolor{blue}{$\bm{53.86 \pm 1.0}$ }\\
\midrule
SVM (source and target) & Amazon+Webcam & $62.91 \pm 1.8$ & $65.82 \pm 1.4$ \\
Late Fusion (Max) & Amazon+Webcam & $65.35 \pm 1.7$ & $58.42 \pm 1.1$ \\
Late Fusion (Lin. Int. Avg) & Amazon+Webcam & $63.23 \pm 1.4$ & $64.29 \pm 1.3$\\
\daume \cite{daume} & Amazon+Webcam & $68.89 \pm 1.9$ & \textcolor{red}{$\bm{72.09 \pm 1.4}$} \\
PMT \cite{aytar-iccv11} & Amazon+Webcam & $64.84 \pm 1.5$ & $65.63 \pm 1.8$ \\
MMDT \cite{hoffman-iclr13} & Amazon+Webcam & $65.47 \pm 1.8$ & $68.10 \pm 1.5$ \\
\midrule
Late Fusion (Lin. Int. Oracle) & Amazon+Webcam & $71.1 \pm 1.7$ & $\bm{72.82 \pm 1.4}$\\
\bottomrule
\end{tabular}

\caption{Amazon$\rightarrow$Webcam adaptation experiment. We show here
  multiclass accuracy on the target domain test set for both supervised and
  unsupervised adaptation experiments across the two fully connected layer
  features (similar to \cite{deeplearning-arxiv-2013}, but with one labeled
  target example). The best performing unsupervised adaptation algorithms are
  shown in blue and the best performing supervised adaptation algorithms are
  shown in red.}

\label{tab:fc6and7_amazon}
\end{table*}

\subsection{Effect of Source Domain Size}
Previous studies considered source domains from the Office dataset. In this section, we ask what happens when an orders-of-magnitute larger source dataset is used.

For completeness we begin by evaluating Amazon as a source domain. 
Preliminary results on this setting are reported in~\cite{deeplearning-arxiv-2013}, here 
we extend the comparison here by
presenting the results with more adaptation algorithms and more complete
evaluation of hyperparameter settings. Table~\ref{tab:fc6and7_amazon} presents
multiclass accuracies for each algorithm using either layer 6 or 7 from the deep
network, which corresponds to the output from each of the fully connected layers.

An SVM trained using only Amazon data achieves 78.6\% in-domain accuracy (tested on the same domain) when using the DeCAF$_6$ feature and 80.2\% in-domain accuracy when using the DeCAF$_7$ feature. These numbers are significantly higher than the performance of the same classifier on Webcam test data, indicating that even with the DeCAF features, there is a still a domain shift between the Amazon and Webcam datasets. 

Next, we consider an unsupervised adaptation setting where no labeled examples are available from the target dataset. In this scenario, we apply two state-of-the-art unsupervised adaptation methods, GFK~\cite{gong-cvpr12} and SA~\cite{sa}. 
Both of these methods make use of a subspace dimensionality hyperparameter.
We show the results using a 100-dimensional subspace and leave the discussion of setting this parameter until Section~\ref{sec:analysis}. For this shift the adaptation algorithms increase performance when using the layer 6 feature, but offer no additional improvement when using the layer 7 feature. 

We finally assume that a single example per category is available in the target domain.
As the bottom rows of Table~\ref{tab:fc6and7_amazon} show, supervised adaptation algorithms are able to provide significant improvement regardless of the feature space chosen, even in the one-shot scenario.
For this experiment we noticed that using the second fully connected layer (DeCAF$_7$) was a stronger overall feature in general.

\subsection{Adapting with a Large Scale Source Domain}

We next address one of the main questions of this paper: Is there still a domain shift when using a large source dataset such as ImageNet? To begin to answer this question we follow the same experimental paradigm as the previous experiment, but use ImageNet as our source dataset. 
The results are shown in Table~\ref{tab:fc6and7_imagenet}.
\begin{table*}[t]
\centering
\begin{tabular}{lccc}
\toprule
Adaptation Method & Training Data & DeCAF$_6$ & DeCAF$_7$ \\
\midrule
SVM (source only) & ImageNet & $53.51 \pm 1.1$ & $59.15 \pm 1.1$ \\
SVM (target only) & Webcam & $62.28 \pm 1.8$ & $64.97 \pm 1.8$ \\
\midrule
GFK \cite{gong-cvpr12} & ImageNet & $65.16 \pm 1.1$ & \textcolor{blue}{$\bm{67.97 \pm 1.4 }$}\\
SA \cite{sa} & ImageNet & $59.30 \pm 1.4$ & $66.08 \pm 1.4$ \\
\midrule
SVM (source and target) & ImageNet+Webcam & $56.68 \pm 1.2$ & $66.93 \pm 1.3$ \\
Late Fusion (Max) & ImageNet+Webcam & $59.59 \pm 1.3$ & $68.86 \pm 1.2$ \\
Late Fusion (Lin. Int. Avg) & ImageNet+Webcam & $60.64\pm 1.3$& $66.45 \pm 1.1$\\
\daume \cite{daume} & ImageNet+Webcam & $59.21 \pm 1.7$ & \textcolor{red}{$\bm{71.39 \pm 1.5}$} \\
PMT \cite{aytar-iccv11} & ImageNet+Webcam & $66.30 \pm 2.1$ & $69.81 \pm 1.8$ \\
MMDT \cite{hoffman-iclr13} & ImageNet+Webcam & $59.21 \pm 1.3$ & $67.75 \pm 1.4$ \\
\midrule
Late Fusion (Lin. Int. Oracle) & ImageNet+Webcam & $71.65 \pm 2.0$ & {$\bm{76.76 \pm 1.3}$} \\
\bottomrule
\end{tabular}
\caption{ImageNet$\rightarrow$Webcam adaptation experiment. Comparison of unsupervised and supervised adaptation algorithms on the ImageNet to Webcam domain shift. Results are computed using the outputs of each of the fully connected layers as features. The best supervised adaptation performance is indicated in red and the best unsupervised adaptation performance is highlighted in blue.}
\label{tab:fc6and7_imagenet}
\end{table*}

Again, we first verify that the source only SVM achieves higher performance when tested on in-domain data than on Webcam data. 
Indeed, for the 16 overlapping labels, the source SVM produces 62.50\% accuracy on ImageNet data using DeCAF$_6$ features and 74.50\% accuracy when using DeCAF$_7$ features. 
Compare this to the 54\% and 59\% for Webcam evaluation and a dataset bias is still clearly evident.

Note that when using ImageNet as a source domain, overall performance of all algorithms improves. In addition, unsupervised adaptation approaches are more effective than for the smaller source domain experiment. 


\subsection{Adapting a Pre-trained Classifier to a New Label Set }
\begin{table*}
\centering
\begin{tabular}{lccc}
\toprule
Adaptation Method & Training Data & Source=ImageNet & Source=Amazon \\
\midrule
SVM (source only) & Source & $66.23 \pm 0.8$ & $53.23 \pm 1.6$ \\
SVM (target only) & Webcam & $63.13 \pm 1.9$ & $63.13 \pm 1.9$ \\
\midrule
GFK \cite{gong-cvpr12} & Source & \textcolor{blue}{$\bm{68.73 \pm 1.1}$} & $54.56 \pm 1.2$ \\
SA \cite{sa} & Source & $66.08 \pm 1.1$ & $55.98 \pm 1.0$ \\
\midrule
SVM (source and target) & Source+Webcam & $75.13 \pm 1.1$ & $63.20 \pm 1.7$ \\
Late Fusion (Max) & Source+Webcam & $71.77 \pm 1.4$ & $62.25 \pm 0.8$ \\
Late Fusion (LinInt Avg) & Source+Webcam & $70.56 \pm 1.2$& $64.56 \pm 1.3$ \\
\daume \cite{daume} & Source+Webcam & \textcolor{red}{$\bm{77.15 \pm 1.1}$} & $70.51 \pm 1.7$ \\
PMT \cite{aytar-iccv11} & Source+Webcam & $70.28 \pm 1.8$ & $66.77 \pm 2.1$ \\
MMDT \cite{hoffman-iclr13} & Source+Webcam & $73.96 \pm 1.2$ & $66.23 \pm 1.4$ \\
\midrule
Late Fusion (Lin. Int. Oracle) & Source+Webcam & {$\bm{76.61 \pm 1.5}$} & $71.49 \pm 1.3$ \\
\bottomrule
\end{tabular}
\caption{ImageNet$\rightarrow$Webcam and Amazon$\rightarrow$Webcam adaptation
  experiments using DeCAF$_8$, the label activations of the CNN trained on the
  full ImageNet data. Again, we compare multiclass accuracy of various
  unsupervised and supervised adaptation methods. The best performing
  unsupervised adaptation algorithm is shown in blue and the best performing
  supervised adaptation algorithms are shown in red.}
\label{tab:fc8}
\end{table*}

DeCAF$_8$ differs from the other DeCAF features in that it constitutes the 1000
activations corresponding to the 1000 labels in the ImageNet classification
task. In the CNN proposed by~\cite{supervision}, these activations are fed into
a softmax unit to compute the label probabilities. We instead experiment with
using the DeCAF$_8$ activations directly as a feature representation, which is
akin to training another classifier using the output of the 1000-way CNN
classifier.

Table~\ref{tab:fc8} shows results for various adaptation techniques using both
ImageNet and Amazon as source domains. We use the same setup as before, but
instead use DeCAF$_8$ as the feature representation. 
The ImageNet results are uniformly better with DeCAF$_8$ than with DeCAF$_6$ or
DeCAF$_7$, likely due to the fact that DeCAF$_8$ was explicitly
trained on ImageNet data to effectively discriminate between ImageNet categories.
Because it can more
effectively classify images from the source domain, it is able to better adapt
from the source domain to the target domain.

However, we see a negligible difference in performance for Amazon, with
performance actually decreasing with respect to DeCAF$_7$ for certain adaptation
methods. We believe this is because the final activation vector is too specific
to the 1000-way ImageNet task, and that DeCAF$_7$ provides a more general
representation that is better suited to the Amazon domain. This, in turn,
results in improved adaptation.
In general, however, the difference between the
various DeCAF representations with Amazon as a source are small enough to be
insignificant.

\subsection{Analysis and Practical Considerations}
\label{sec:analysis}
Our adaptation experiments show that, despite its large size, even ImageNet is not large enough to cover all domains, and
that traditional domain adaptation methods go a long way in increasing performance and mitigating the effects of this shift.
Depending on the characteristics of the problem at hand, our results suggest different methods may be most suitable.

\begin{figure}
\centering
\subfigure[Late Fusion with Linear Interpolation]{
\includegraphics[height=.4\linewidth]{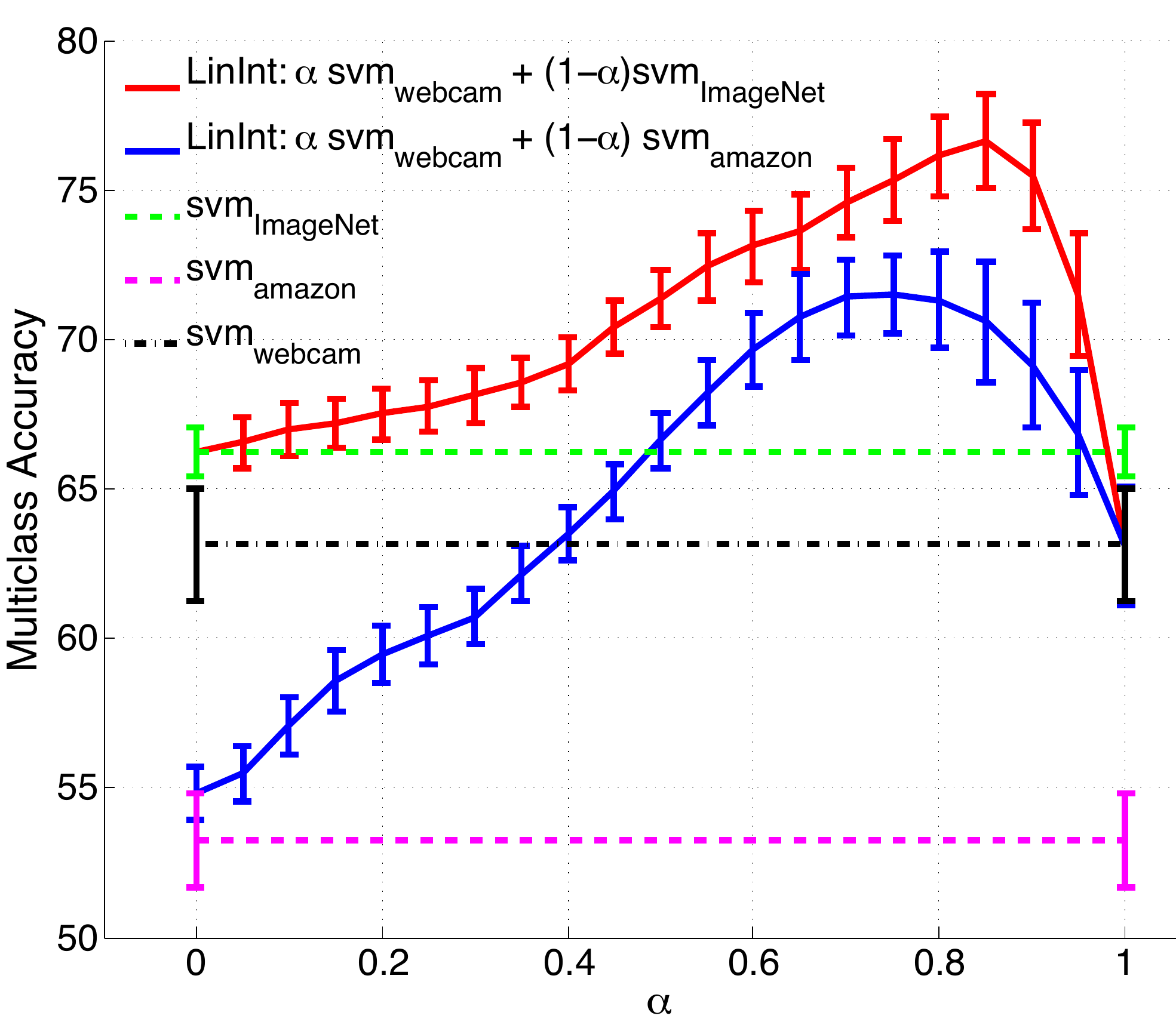}
\label{fig:linint-eval}
}
\subfigure[Unsupervised Methods]{
\includegraphics[height=.4\linewidth]{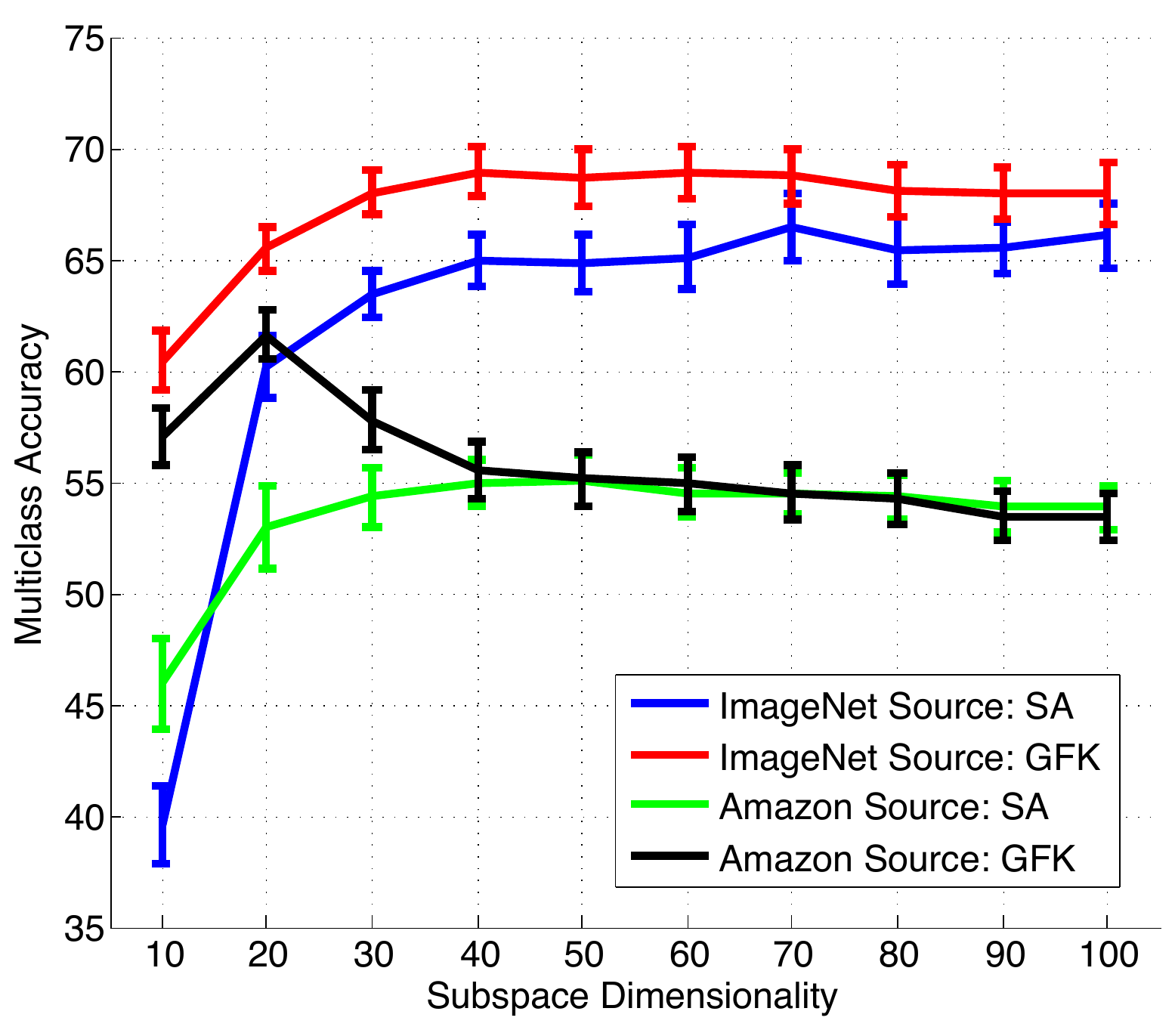}
\label{fig:sagfk-eval}
}
\caption{Evaluation of hyperparameters for domain adaptation methods. (a) Analysis of the combination hyperparameter $\alpha$ for Late Fusion with linear interpolation. (b) Analysis of the subspace dimensionality for the unsupervised adaptation algorithms}
\label{fig:hyperparam-eval}
\end{figure}

If no labels exist in the target domain, then there are unsupervised adaptation algorithms that are easy to use and fast to compute at adaptation time, yet still achieve increased performance over source-only methods. 
For this scenario, we experimented with two subspace alignment based methods that both require setting a parameter that indicates the dimensionality of the input subspaces. 
Figure~\ref{fig:sagfk-eval} shows the effect that changing the subspace dimensionality has on the overall method performance. 
In general, we noticed that these methods were not particularly sensitive to this parameter so long as the dimensionality remains larger than the number of categories in our label set.
Below this threshold, the subspace is less likely to capture all important discriminative information needed for classification.

In the case where we have a large source dataset and a limited number of labeled target examples, it may be preferable to compute source classifier parameters in advance, then examine only the source parameters and the target data at adaptation time.
Examples of these kinds of methods are Late Fusion and PMT.
These methods are unaffected by the number of data points in the source domain at adaptation time, and can thus be applied quickly.
In our experiments, we found that a properly tuned Late Fusion classifier with linear interpolation was the fastest and most effective approach.
Figure~\ref{fig:linint-eval} shows the performance of linear interpolation Late Fusion as we vary the hyperparameter $\alpha$.
Although the method is sensitive to $\alpha$, we found that for both source domains, the basic strategy of setting $\alpha$ around $0.8$ provides a close approximation to optimal performance. 
This setting can be interpreted as trusting the target classifier more than the source, but not so much as to completely discount the information available from the source classifier. In each table we report both the performance of linear interpolation both averaged across hyper parameter settings $\alpha \in [0,1]$ as well as the performance of linear interpolation with the best possible setting of $\alpha$ per experiment -- this is denoted as ``Oracle" performance.

If there are no computational constraints and there are very few labels in
the target domain, the best-performing method seems to be the ``frustratingly
easy'' approach originally proposed by \daume~\cite{daume} and applied again for
deep models in \cite{ref:dlid}.

Finally, we found that feature representation can have a significant impact on
adaptation performance. Our results show that ImageNet as source performs best
with the DeCAF$_8$ representation, whereas Amazon as source performs best with
the DeCAF$_7$ representation. This, combined with our intuition, seems to
indicate that for adaptation from source domains other than ImageNet, an
intermediate representation other than DeCAF$_8$ is more powerful for
adaptation, whereas ImageNet classification works best with the full representation that
was trained on it.

\section{Conclusion}
In this paper, we presented the first evaluation of domain adaptation from a
large-scale source dataset with deep features. We demonstrated that, although using ImageNet as a
source domain generalizes better than other smaller source domains, there is
still a domain shift when adapting to other visual domains.

Our experimental results show that deep adaptation methods can go a long
way in mitigating the effects of this domain shift. Based on our results, we
also provided a set of practical recommendations for choosing a feature
representation and adaptation method accounting for constraints on runtime and
accuracy.

There are a number of interesting directions to take given our results. First we
notice that though DeCAF$_8$ is the strongest feature to use for learning a
classifier on ImageNet data, DeCAF$_7$ is actually a better feature to use with
the Amazon source domain and the Webcam target domain. This could lead to a
hybrid approach where one uses different feature representations for the various
domains and produces a combined adapted model. Another interesting direction
that should be explored is to integrate the adaption algorithms into the deep
models explicitly and even allow for feedback between the two stages. Current
deep models although allow information flow between the final classifier and the
representation learning architecture. We feel that the next step is to have a
separate task specific adaptable layer that does not simply learn a new final
layer, but instead learns a separate, but equivalent final layer, that is
regularized by the final layer learned on the source dataset.

This future work is a natural extension of the result we have shown in this
paper: that pre-trained deep representations with large source domains can be
effectively adapted to new target domains using only shallow, linear adaptation
methods, and that in cases where the target data is limited, this approach is
the best way to mitigate dataset bias.

\small{
\bibliographystyle{plain}
\bibliography{main}
}

\end{document}